\documentclass[runningheads]{llncs}

\usepackage{eccv}

\usepackage{eccvabbrv}

\usepackage{graphicx}
\usepackage{booktabs}
\usepackage{multirow}
\usepackage{pifont}
\usepackage{bm}
\usepackage{wrapfig}
\usepackage[table]{xcolor}

\usepackage{marvosym}

\usepackage[accsupp]{axessibility}  

\usepackage{hyperref}

\usepackage{orcidlink}

\begin{document}

\title{From Reasoning to Pixels: Grounded Medical Multimodal LLMs for VQA and Segmentation} 
\titlerunning{MedREAL: From Reasoning to Pixels}

\author{Haowen Gu\inst{1,2}\orcidlink{0009-0003-5418-0596} \and
Gensheng Pei\inst{3}\orcidlink{0000-0002-7677-7487} \and
Junzhu Mao\inst{1,2}\orcidlink{0009-0005-0831-9272} \and
Qiong Wang\inst{1,2}\orcidlink{0000-0003-4193-0960} \and \\
Mingwu Ren\inst{1,2\textsuperscript{(\Letter)}}\orcidlink{0000-0001-5576-3281} \and
Yazhou Yao\inst{1,2\textsuperscript{(\Letter)}}\orcidlink{0000-0002-0337-9410}}

\authorrunning{H.~Gu et al.}

\institute{Nanjing University of Science and Technology, Nanjing, China\and
State Key Laboratory of Intelligent Manufacturing of Advanced Construction Machinery, Nanjing, China \and
Department of Electrical and Computer Engineering, Sungkyunkwan University, Suwon, Korea \\
\email{yazhou.yao@njust.edu.cn, renmingwu@mail.njust.edu.cn} \\
\url{https://github.com/NUST-Machine-Intelligence-Laboratory/MedREAL}}

\maketitle

\begin{abstract}
Although Multimodal Large Language Models (MLLMs) have demonstrated impressive performance in Medical Visual Question Answering (Med-VQA), their reliance on global image features often lacks precise pixel-level grounding, thereby limiting clinical trustworthiness. To bridge the semantic gap between high-level clinical reasoning and spatial localization, we propose \textsc{\textsc{MedREAL}} (\textbf{Med}ical \textbf{RE}asoning-driven \textbf{A}nswering and \textbf{L}ocalization), a unified framework that seamlessly aligns linguistic reasoning with spatial grounding. Specifically, \textsc{MedREAL} introduces \textbf{S}eg \textbf{A}nchored \textbf{R}easoning \textbf{P}ooling (SARP) to distill task-relevant semantic evidence directly from \texttt{[SEG]} tokens within the MLLM's hidden states. Furthermore, a \textbf{R}easoning-to-\textbf{V}isual (R2V) fusion mechanism is proposed to effectively inject these reasoning-aware features into a segmentation pipeline for accurate mask decoding. To facilitate this paradigm, we construct MedRAVS-13K, a comprehensive dataset comprising 13,824 expertly validated samples across four diverse imaging modalities. Extensive experiments demonstrate that \textsc{MedREAL} significantly outperforms state-of-the-arts, achieving 68.49\% gIoU and 70.47\% cIoU on benchmark evaluations. By generating evidence masks that are strictly consistent with textual diagnoses, \textsc{MedREAL} provides a robust, interpretable framework for reasoning-driven medical image analysis.
\end{abstract}

\section{Introduction}
\label{sec:intro}

Medical visual question answering (Med-VQA) aims to answer clinically relevant questions about medical images, requiring models to jointly understand visual content and perform domain-specific reasoning. Beyond answer prediction, clinical practice often demands explicit localization of the visual evidence that supports a decision, such as identifying a lesion, anatomical structure, or pathological region. This requirement highlights a critical limitation of existing Med-VQA systems: while recent multimodal large language models (MLLMs)\cite{bai2023qwen,liu2023llava,liu2024llavanext,achiam2023gpt4,comanici2025gemini,wei2025deepseekocr} demonstrate strong reasoning capabilities, they typically provide answers without grounding them in pixel-level evidence, thereby reducing interpretability and limiting clinical trust.

\begin{figure}[t]
\centering
\includegraphics[width=\linewidth]{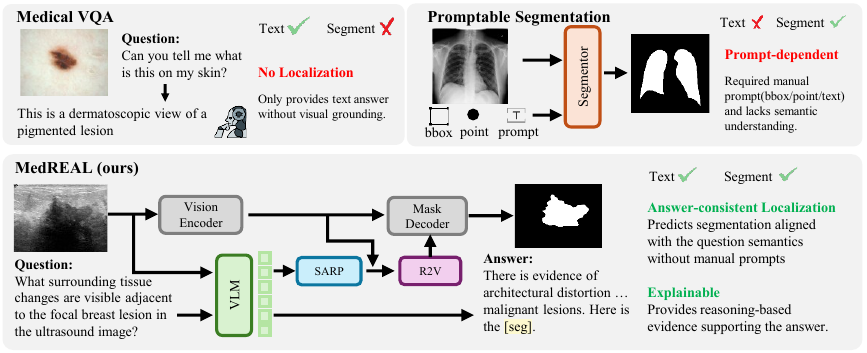}
\caption{Comparison between Medical VQA, promptable segmentation, and the proposed \textsc{MedREAL} framework. \textsc{MedREAL} introduces an explicit evidence token to extract reasoning-aligned features, which are fused with visual context to produce answer-consistent evidence localization.}
\label{fig:figure1}
\end{figure}

Most medical image segmentation methods\cite{zhang2025semantic,lv2025test,FanWen_AnatomyAware_MICCAI2025,ma2024segment,wei2024medsam,butoi2023universeg} demonstrate strong capability in spatial localization, yet remain limited in high-level semantic understanding. Although recent promptable foundation models such as SAM3\cite{carion2025sam} and its medical variant MedSAM3\cite{liu2025medsam3} enable concept segmentation from text-conditioned inputs, their semantic modeling remains shallow and largely relies on lexical alignment rather than complex clinical reasoning. As a result, these approaches struggle to capture the multi-step diagnostic logic required in real-world scenarios, where segmentation should be guided by question-driven semantic evidence rather than generic visual prompts. This creates a semantic disconnect between high-level diagnostic logic and pixel-level spatial grounding. Overcoming this challenge is essential for building integrated systems capable of joint medical answering and localization, ensuring that every diagnostic response is backed by spatially consistent visual justification.

Simply concatenating MLLMs with segmentation models, however, is suboptimal. Although MLLM hidden states contain rich multimodal information, they are often dominated by global context and broad linguistic patterns, making it difficult to isolate the specific visual features that serve as reasoning evidence. Furthermore, directly feeding text embeddings into a segmentation model leads to poor spatial alignment, as the connection between token-level reasoning and pixel-level localization is not explicitly modeled. As illustrated in Figure \ref{fig:figure1}, an effective solution must bridge this gap by explicitly identifying reasoning-related evidence within MLLM representations and transforming it into spatially meaningful signals to guide the segmentation process.

To address these challenges, we propose \textsc{\textsc{MedREAL}} (\textbf{Med}ical \textbf{RE}asoning-driven \textbf{A}nswering and \textbf{L}ocalization), a unified framework that aligns linguistic reasoning with pixel-level localization within a reasoning-aware latent space. Specifically, we embed an explicit evidence token, \texttt{[SEG]}, into the Med-VQA generation sequence, utilizing its hidden representation as a reasoning anchor to distill the critical semantic evidence that justifies the predicted answer.
Building upon this design, we introduce the \textbf{S}eg-\textbf{A}nchored \textbf{R}easoning \textbf{P}ooling (SARP) module. SARP leverages the \texttt{[SEG]} representation as a guiding prior to distill reasoning-relevant features from the MLLM's terminal layers. By effectively suppressing irrelevant global semantics, SARP yields a compact, evidence-aligned representation tailored to the target region. Furthermore, we propose a \textbf{R}easoning-to-\textbf{V}isual (R2V) Fusion mechanism, which integrates these distilled reasoning features with the global visual context extracted from the segmentor's image encoder. This operation produces a semantically informed conditioning vector that robustly guides the SAM-based mask decoding stage. With this unified architecture, segmentation is driven not only by visual appearance but also by the model's internal diagnostic reasoning, allowing \textsc{MedREAL} to achieve spatially precise and answer-consistent evidence localization.

Another fundamental challenge in this domain is the scarcity of benchmarks that simultaneously evaluate clinical reasoning and spatial grounding. Current Med-VQA datasets\cite{lau2018dataset,he2020pathvqa,liu2021slake} predominantly prioritize textual accuracy without pixel-level supervision; conversely, existing medical segmentation datasets\cite{al2020dataset,tahir2021covid,codella2019skin,jha2019kvasir,bilic2023lits,riedel2024isles} typically lack reasoning-oriented queries. To enable the training and evaluation of reasoning-driven segmentation, we construct MedRAVS-13K (Medical Reasoning, Answering, and Visual Segmentation), a reasoning-aware benchmark comprising 13,824 expertly curated samples. We augment four heterogeneous medical imaging datasets (\ie, BUSI\cite{al2020dataset}, COVID-QU-Ex\cite{tahir2021covid}, ISIC-2018\cite{codella2019skin}, and Kvasir-SEG\cite{jha2019kvasir}), spanning ultrasound, X-ray, dermoscopy, and endoscopy modalities. We enrich these datasets with clinically oriented question-answer pairs that explicitly refer to specific regions of interest. Each QA pair is matched with a verified pixel-level segmentation mask, forming perfectly aligned image-text-mask triplets. This provides the structured supervision necessary to link high-level diagnostic reasoning with precise spatial localization.

Extensive experiments on the MedRAVS-13K dataset demonstrate that our \textsc{\textsc{MedREAL}} significantly enhances segmentation quality. Compared to existing representative methods, \textsc{MedREAL} achieves state-of-the-art performance with an overall gIoU of 68.49\% and cIoU of 70.47\%. Crucially, by aligning linguistic reasoning with pixel-level grounding, our method ensures that the generated evidence masks are semantically consistent with the predicted clinical answers. This synergy effectively mitigates the ``black-box'' nature of traditional MLLMs, offering substantially improved interpretability and reliability for clinical decision support systems.
In summary, our primary contributions are fourfold:
\begin{itemize}
\item We propose \textsc{MedREAL}, a unified framework that couples Med-VQA reasoning with pixel-level grounding by embedding explicit \texttt{[SEG]} tokens as reasoning anchors to distill semantic evidence from MLLM hidden states.
\item We introduce the SARP module and R2V fusion mechanism to effectively extract reasoning-relevant features and systematically inject them into a SAM-based pipeline for precise mask decoding.
\item We construct MedRAVS-13K, a comprehensive benchmark dataset of 13,824 samples across four diverse imaging modalities, providing a robust foundation of clinical QA pairs and verified pixel-level masks.
\item Extensive evaluations show that \textsc{MedREAL} significantly outperforms existing methods, achieving \textbf{68.49\% gIoU} and \textbf{70.47\% cIoU}, while ensuring high semantic consistency between linguistic reasoning and spatial evidence.
\end{itemize}

\section{Related Work}

\noindent\textbf{Medical Visual Question Answering.}
Medical visual question answering aims to empower models to resolve clinically relevant queries by synthesizing visual content with domain-specific medical knowledge. While early efforts primarily focused on fusion strategies between static image features and linguistic encodings\cite{zhou2018employing,gong2021cross,xu2026gsv2x,gu2026medfg}, the field has recently been transformed by MLLMs. These advanced architectures\cite{li2023llava,tu2024towards,chen2024towards,lai2026med,sun2025jo} leverage massive pretraining to demonstrate sophisticated cross-modal reasoning, enabling them to interpret complex clinical descriptions and generate nuanced natural language responses. However, a critical gap remains as these models predominantly operate on global semantic representations. Despite their linguistic fluency, they typically fail to provide explicit pixel-level grounding for their predictions. This absence of spatial evidence makes it difficult to verify the decision rationale against specific pathological regions, thereby limiting their interpretability and clinical trustworthiness in high-stakes diagnostic environments. To overcome this, \textsc{MedREAL} embeds an \texttt{[SEG]} token as a reasoning anchor to distill local semantic evidence, successfully bridging the gap between linguistic answers and precise spatial grounding.

\noindent\textbf{Medical Image Segmentation.}
Medical image segmentation is a fundamental task in medical image analysis, aiming to accurately delineate lesions or anatomical structures to support diagnosis and treatment planning. Traditional methods\cite{wu2019simultaneous,cao2022swin,huang2020unet,pei2025seeing,Yin2026PCASeg,sun2026bridging} rely on U-Net\cite{ronneberger2015u} and its variants, which perform pixel-wise classification through fully convolutional networks. More recently, promptable segmentation approaches, such as SAM\cite{ma2024segment}, have demonstrated strong generalization across multiple modalities and large-scale datasets, enabling segmentation guided by points, bounding boxes, or textual prompts. However, these methods\cite{wei2024medsam,zhu2024medical,konwer2025enhancing,zhang2024domesticating, jiang2026medical} typically rely solely on visual information and lack language-based reasoning constraints, making it challenging to perform question-driven segmentation. In clinical diagnostic workflows, it is essential for models to provide localized evidence that directly corresponds to a specific query; therefore, the effective integration of linguistic reasoning with spatial segmentation remains an open and critical challenge. Unlike conventional visual-centric models, our \textsc{MedREAL} framework explicitly injects MLLM-derived clinical logic into the segmentation pipeline via the R2V mechanism, ensuring masks are directly driven by diagnostic reasoning.

\noindent\textbf{Reasoning Guided Segmentation.}
Reasoning-guided segmentation aims to incorporate high-level semantic reasoning into pixel-level predictions. Existing studies\cite{lai2024lisa,ren2024pixellm,yan2024visa,liu2025seg,wang2024llm,howlader2026cora,ghezloo2025pathfinder,cai2026iris,cai2026unbiased,yang2026beyond,yang2026efficiency,yang2025changetitans} have largely focused on natural scenes, predominantly relying on a single \texttt{[seg]} token to represent the entire target semantic mask. While this design enables coarse alignment between language and the predicted mask, it represents the target as a holistic concept and overlooks fine-grained evidence that is critical for precise localization. In medical imaging, this line of research is still in its early stages. Some approaches\cite{huang2024seg,liu2025medsam3} attempt to fuse attention maps or intermediate features from question-answering models with segmentation networks to enhance semantic consistency, but they typically rely on global language representations or simple feature concatenation, lacking explicit anchoring mechanisms to localize regions relevant to the question. Moreover, the scarcity of high-quality reasoning-annotated datasets for medical tasks makes joint training and evaluation challenging. \textsc{MedREAL} uniquely addresses these bottlenecks by utilizing the SARP module to extract localized, fine-grained semantic evidence rather than holistic representations, supported by our comprehensive MedRAVS-13K benchmark for rigorous evaluation.

\begin{figure}[t]
    \centering
    \includegraphics[width=\linewidth]{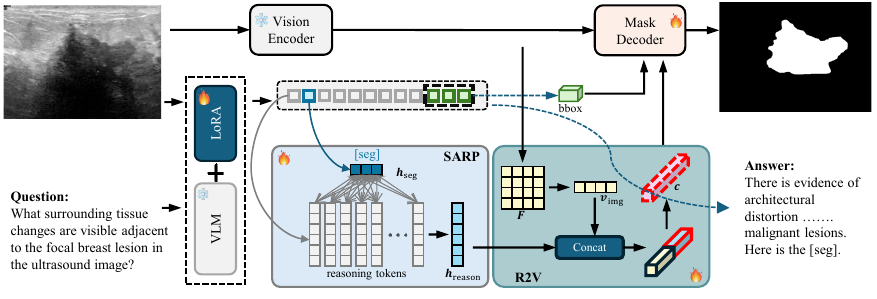}
    \caption{Overview of \textsc{MedREAL}. A medical image and question are processed by a VLM producing answer tokens including [SEG]. SARP distills reasoning-relevant features from [SEG], which are fused with global visual features via R2V Fusion to condition the mask decoder, producing an evidence-aligned segmentation mask guided by both reasoning and visual context.}
    \label{fig:figure2}
\end{figure}

\section{Method}

\subsection{Overall Pipeline}
We propose \textsc{MedREAL}, a unified framework that couples medical visual question answering with pixel-level segmentation by explicitly injecting multimodal reasoning semantics into the localization process. The overall architecture is illustrated in Figure \ref{fig:figure2}.

Unlike standard pipeline models that treat text generation and segmentation as isolated steps, \textsc{MedREAL} unifies them within a reasoning-aware latent space. Given a medical image $\bm{I}$ and a clinical text query $\bm{Q}$, the multimodal large language model (MLLM) autoregressively generates a textual response $\bm{A}$ alongside a sequence of latent hidden states $\bm{H}$. To explicitly bridge language and vision, we prompt the MLLM to output a dedicated \texttt{[SEG]} token when localization is required. The language generation process can be formulated as:
\begin{equation}
    \bm{A}, \bm{H} = \mathcal{F}_{\text{MLLM}}(\bm{I}, \bm{Q})
\end{equation}
where $\bm{H} \in \mathbb{R}^{B \times L \times D}$ encapsulates the dense semantic representations of the generated reasoning process.

Rather than relying solely on generic visual prompts, \textsc{MedREAL} leverages this textual reasoning to guide the spatial decoding. We introduce the Seg-Anchored Reasoning Pooling (\textbf{SARP}) module to extract a compact reasoning representation $\bm{h}_{\text{reason}}$ from $\bm{H}$. Subsequently, the Reasoning-to-Visual (\textbf{R2V}) Fusion module dynamically integrates $\bm{h}_{\text{reason}}$ with the visual feature maps $\bm{F}$ extracted by the segmentor's image encoder. This yields a semantically informed prompt $\bm{c}$, which is fed into the mask decoder $\mathcal{D}_{\text{mask}}$ to predict the final evidence mask $\bm{M}$:
\begin{equation}
    \bm{M} = \mathcal{D}_{\text{mask}}(\bm{F}, \bm{c}).
\end{equation}

By formulating the pipeline in this cohesive manner, the segmentation mask is strictly conditioned on the diagnostic logic formulated by the MLLM, ensuring semantic alignment between the textual answer and the localized visual evidence.

\subsection{Seg-Anchored Reasoning Pooling}
While the \texttt{[SEG]} token acts as a trigger for the segmentation task, extracting features exclusively from its corresponding hidden state is suboptimal. Clinical reasoning is inherently distributed across the preceding linguistic sequence; a single token lacks the capacity to encapsulate complex diagnostic multi-step logic. To address this, we propose the SARP module, which utilizes the \texttt{[SEG]} token as an attention anchor to aggregate reasoning-relevant semantics from the broader context.

Let $\bm{H} \in \mathbb{R}^{B \times L \times D}$ denote the terminal-layer hidden states of the MLLM. We first isolate the representation of the \texttt{[SEG]} token using a binary positional mask $\bm{m}^s \in \{0, 1\}^L$. To ensure robustness against potential sequence shifts, the segmentation anchor $\bm{h}_{\text{seg}} \in \mathbb{R}^{D}$ is computed via masked average pooling:
\begin{equation}
    \bm{h}_{\text{seg}} = \frac{\sum_{i=1}^{L} m_{i}^{s} \bm{H}_{i}}{\sum_{i=1}^{L} m_{i}^{s} + \epsilon},
\end{equation}
where $\epsilon$ is a small constant to prevent zero division. This anchor explicitly captures the model's semantic state at the exact moment spatial grounding is invoked.
To adaptively filter out irrelevant linguistic noise (\eg, grammatical structures or generic clinical boilerplate) and concentrate on specific diagnostic evidence, we project $\bm{h}_{\text{seg}}$ into a query vector $\bm{q}$. Concurrently, the preceding reasoning sequence within $\bm{H}$ is linearly mapped into key $\bm{K}$ and value $\bm{V}$ matrices. An attention-based pooling mechanism is then applied to measure the relevance of each reasoning token to the segmentation trigger:

\begin{equation}
    \bm{\alpha} = \text{Softmax}\left(\frac{\bm{q} \bm{K}^\top}{\sqrt{d_k}}\right),
\end{equation}
where $d_k$ is the scaling factor based on the hidden dimension. The final reasoning representation $\bm{h}_{\text{reason}}$ is derived through a weighted summation:
\begin{equation}
    \bm{h}_{\text{reason}} = \bm{\alpha} \bm{V}.
\end{equation}
This operation distills a compact, high-density semantic vector that explicitly encodes the clinical rationale necessitating the segmentation, providing a highly informative prior for the subsequent localization phase.

\subsection{Reasoning-to-Visual Fusion}
Although $\bm{h}_{\text{reason}}$ is semantically rich, it exists entirely within the linguistic latent space and lacks the spatial inductive biases required for precise pixel-level mask generation. To translate abstract diagnostic logic into an explicit spatial guidance signal, we introduce the R2V Fusion module, which bridges the extracted reasoning with global anatomical context.

Given the multi-scale visual feature map $\bm{F} \in \mathbb{R}^{B \times C \times H' \times W'}$ from the segmentor's image encoder, we first apply global average pooling to collapse the spatial dimensions. A linear projection followed by a non-linear activation $\sigma$ maps this representation into a global visual context vector $\bm{v}_{\text{img}} \in \mathbb{R}^{B \times D_c}$:
\begin{equation}
    \bm{v}_{\text{img}} = \sigma\Big(\text{Linear}\big(\text{AvgPool}(\bm{F})\big)\Big).
\end{equation}

This vector provides a structural anatomical prior. To enable deep multimodal interaction, we concatenate $\bm{v}_{\text{img}}$ with the reasoning representation $\bm{h}_{\text{reason}}$ along the channel dimension. The fused tensor is then projected through a two-layer multi-layer perceptron (MLP) parameterized by $\phi$:
\begin{equation}
    \bm{c} = \phi\big([\bm{v}_{\text{img}}, \bm{h}_{\text{reason}}]\big),
\end{equation}
where $[\cdot, \cdot]$ denotes the concatenation operation. The resulting conditional vector $\bm{c}$ fundamentally alters the segmentation paradigm: rather than relying on geometric prompts (\eg, points or boxes), $\bm{c}$ acts as a semantically-informed pseudo-prompt for the SAM-based mask decoder. This ensures the decoded mask strictly adheres to the diagnostic evidence articulated by the MLLM.

\subsection{Optimization Objective}
The entire \textsc{MedREAL} framework is trained end-to-end, jointly optimizing both the text generation quality and the spatial segmentation accuracy. The total loss $\mathcal{L}$ is formulated as a weighted combination:
\begin{equation}
    \mathcal{L} = \lambda_{\text{txt}} \mathcal{L}_{\text{CE}}(\hat{\bm{Y}}_{\text{txt}}, \bm{Y}_{\text{txt}}) + \lambda_{\text{mask}} \Big[ \lambda_{\text{bce}} \mathcal{L}_{\text{BCE}}(\hat{\bm{M}}, \bm{M}) + \lambda_{\text{dice}} \mathcal{L}_{\text{DICE}}(\hat{\bm{M}}, \bm{M}) \Big].
\end{equation}
Here, $\mathcal{L}_{\text{CE}}$ represents the standard auto-regressive cross-entropy loss applied to the textual response. For the spatial localization, the mask loss is computed as a weighted sum of the per-pixel Binary Cross-Entropy ($\mathcal{L}_{\text{BCE}}$) and the Dice loss ($\mathcal{L}_{\text{DICE}}$), which handles class imbalance in medical lesions. This composite objective effectively bridges the MLLM and the visual segmentor, forcing the shared latent space to align linguistic logic with pixel-level boundaries.

\section{Dataset}
\label{sec:dataset}

\subsection{Data Source}

\begin{wrapfigure}{r}{0.52\textwidth}
\centering
\includegraphics[width=\linewidth]{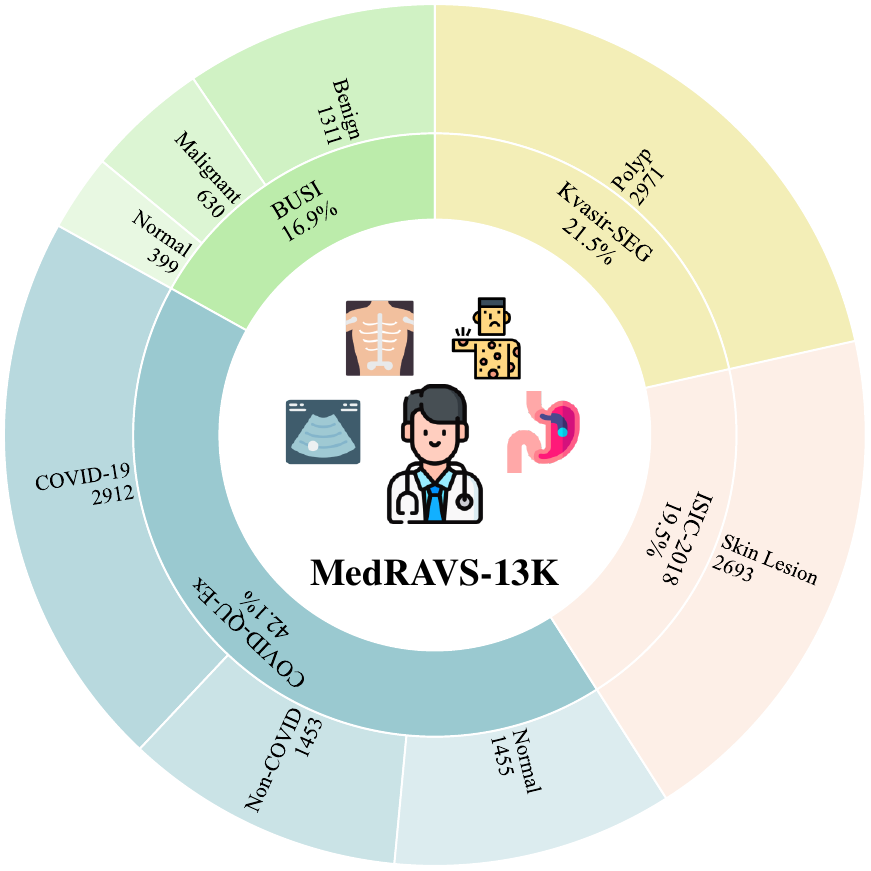}
\caption{Overview of MedRAVS-13K.}
\label{fig:dataset_overview}
\end{wrapfigure}

As summarized in Figure \ref{fig:dataset_overview}, we construct \textbf{MedRAVS-13K} (Medical Reasoning, Answering, and Visual Segmentation) to rigorously evaluate the \textsc{MedREAL} framework. We integrate four diverse publicly available sources: BUSI\cite{al2020dataset}, COVID-QU-Ex\cite{tahir2021covid}, ISIC-2018\cite{codella2019skin}, and Kvasir-SEG\cite{jha2019kvasir}. Unlike conventional benchmarks that treat text and masks in isolation, MedRAVS-13K explicitly aligns fine-grained clinical question-answering with precise visual segmentation masks. This design enables a dual assessment of sophisticated diagnostic reasoning and localized evidence generation.

The dataset spans multiple modalities, \eg, ultrasound, X-ray, dermoscopy, and endoscopy, covering critical organs such as the breast, lung, skin, and colon. These sources exhibit substantial heterogeneity in imaging physics and lesion morphology, ranging from low-contrast sonographic boundaries to intricate endoscopic mucosal textures. Such multi-modality composition provides a robust and challenging testbed for evaluating reasoning-driven segmentation across clinically distinct scenarios.

\subsection{Data Generation and Curation}

\begin{figure}[t]
\centering
\includegraphics[width=1.0\textwidth]{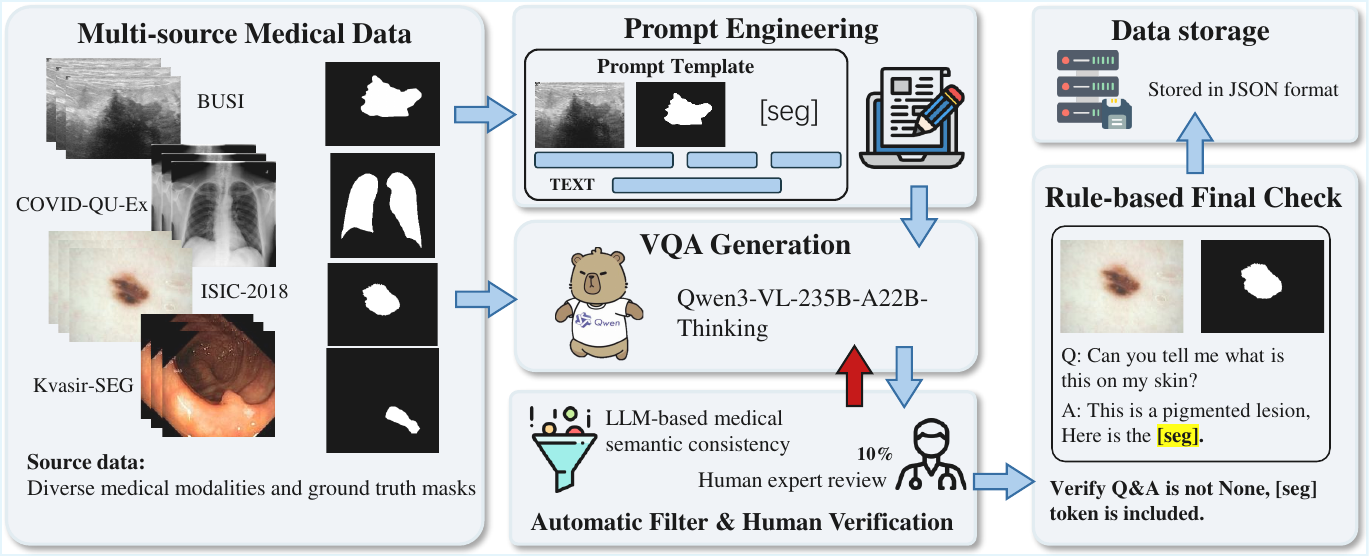}
\caption{Dataset Generation Pipeline of MedRAVS-13K. The process illustrates the integration of raw images and masks to generate reasoning-aware text, followed by rigorous quality assurance.}
\label{fig:dataset_gen}
\end{figure}

Figure \ref{fig:dataset_gen} details the multi-stage generation and curation pipeline designed to align clinical reasoning with pixel-level annotations. For each source dataset, we develop modality- and organ-specific prompts that incorporate the raw medical images alongside their ground-truth masks. We leverage the advanced reasoning capabilities of Qwen3-VL-235B-A22B-Thinking to generate clinically accurate question-answering (QA) pairs. Crucially, a \texttt{[SEG]} token is explicitly embedded within each generated sequence to denote the region of interest. This mechanism ensures that the textual rationale is strictly grounded in the annotated anatomy or lesion, compelling the model to bridge abstract diagnostic logic with concrete visual evidence.

To guarantee data fidelity, we implement a two-tier quality assurance protocol. First, an automated filtering stage employs a powerful large language model to evaluate the medical plausibility, semantic consistency, and visual relevance of each generated pair, discarding low-quality outputs. Second, we perform meticulous manual verification on a 10\% subset of the data to confirm clinical accuracy and verify the spatial alignment between the textual queries and the segmentation masks. To mitigate potential biases arising from class and modality imbalances, we apply a controlled sampling strategy across the pathological categories during the final split construction, ensuring a representative benchmark.
Ultimately, MedRAVS-13K comprises over 13,000 expertly curated samples. Each instance seamlessly integrates a medical image, a \texttt{[SEG]}-augmented QA pair, and a corresponding pixel-level mask. These queries encompass diverse diagnostic intents, including polar (yes/no), descriptive, and localization-focused questions. Because the masks exhibit substantial variation in scale and morphology across the four modalities, the dataset presents highly challenging visual grounding scenarios that demand the joint optimization of reasoning, segmentation, and answer prediction.

\section{Experiments}

\subsection{Experiment Setup}
\textbf{Datasets.} To evaluate the performance of \textsc{MedREAL}, all experiments are conducted on the MedRAVS-13K dataset as described in Sec. \ref{sec:dataset}. For comparative analysis involving baseline models that lack inherent reasoning capabilities, we utilize standard category labels to implement the segmentation task. Detailed specifications regarding the prompting mechanisms for various architectures are provided in the subsequent sections.

\noindent\textbf{Evaluation Metrics.} Following established protocols in reasoning segmentation, we adopt gIoU and cIoU as primary metrics. gIoU is defined as the arithmetic mean of per image IoU scores, while cIoU represents the ratio of total intersection to total union across the dataset. Since cIoU is heavily biased toward large area objects and prone to instability, gIoU is preferred in medical contexts where lesion scales vary significantly, as it provides a more balanced measure of segmentation quality.

\noindent\textbf{Implementation Details.} Our model is trained on two NVIDIA 48G A6000 GPUs using the DeepSpeed \cite{rasley2020deepspeed} engine to enable memory-efficient optimization and stable large scale multimodal training. We adopt the AdamW \cite{loshchilov2019decoupledweightdecayregularization} optimizer with an initial learning rate of $2\times10^{-4}$ and apply weight decay for regularization. Unless otherwise specified, we adopt Qwen3-VL-2B-Instruct \cite{bai2025qwen3vl} as the MLLM and SAM \cite{kirillov2023segment} as the segmentor for all experiments. Following the optimization strategy in LISA \cite{lai2024lisa}, we set the loss weights $\lambda_\text{txt}$ and $\lambda_\text{mask}$ to 1.0. For the mask supervision components, $\lambda_\text{bce}$ and $\lambda_\text{dice}$ are assigned values of 2.0 and 0.5.

\begin{table}[t]
    \centering
    \caption{Referring phrases used for different datasets and semantic classes. These ground-truth derived templates serve as explicit prompts for non-reasoning baselines.}
    \label{tab:referring_phrases}
    \renewcommand{\arraystretch}{1.0}
    \setlength{\tabcolsep}{1pt}
    \resizebox{\linewidth}{!}{
    \begin{tabular}{@{} l l c c l @{}}
        \toprule
        \textbf{Dataset} & \textbf{Class} & \textbf{Train Samples} & \textbf{Val Samples} & \textbf{Referring Phrase} \\
        \midrule
        
        \multirow{3}{*}{BUSI\cite{al2020dataset}} 
        & Benign     & 915 & 396 & benign breast mass \\
        & Malignant  & 441 & 189 & malignant breast tumor \\
        & Normal     & 276 & 123 & normal breast tissue \\
        \addlinespace
        
        \multirow{3}{*}{COVID-QU-Ex\cite{tahir2021covid}}
        & COVID-19   & 2330 & 582 & COVID-19 lung infection \\
        & Non-COVID  & 1162 & 291 & non-COVID lung infection \\
        & Normal     & 1164 & 291 & normal lung region \\
        \addlinespace
        
        ISIC-2018\cite{codella2019skin} 
        & Skin Lesion & 2593 & 100 & skin lesion region \\
        \addlinespace
        
        Kvasir-SEG\cite{jha2019kvasir} 
        & Polyp       & 2615 & 356 & polyp region \\
        
        \bottomrule
    \end{tabular}}
\end{table}

\subsection{Comparison with State-of-the-Art Methods}

To evaluate \textsc{MedREAL}, we benchmark it against three distinct paradigms: referring-based segmentation (\ie, OVSeg\cite{liang2023open}, SAM3\cite{carion2025sam}, MedSAM3\cite{liu2025medsam3}), agent-based iterative systems (\ie, SAM3-Agent\cite{carion2025sam}, MedSAM3-Agent\cite{liu2025medsam3}), and end-to-end reasoning segmentation (\ie, LISA\cite{lai2024lisa}). 

Because referring-based purely structural models lack intrinsic reasoning capabilities, they cannot process the complex VQA queries in MedRAVS-13K directly. To accommodate them and establish an upper-bound for structural decoding, we provide explicit, oracle-like text prompts derived from the ground-truth category of each query, with detailed templates summarized in Table \ref{tab:referring_phrases}. For the agent-based approaches, we deploy Qwen3-VL-8B-Thinking as the reasoning engine to autonomously generate prompts for the segmentors (capped at 20 rounds). Finally, LISA is evaluated as the primary reasoning-segmentation baseline, utilizing LLaVA-7B\cite{li2023llava} as its multimodal backbone.

\begin{table*}[t]
    \centering
    \caption{Comparison of gIoU and cIoU across different datasets. Best results are highlighted in \textbf{bold}, and second-best are \underline{underlined}.}
    \label{tab:seg_results}
    \renewcommand{\arraystretch}{1.0}
    \setlength{\tabcolsep}{2pt}
    \resizebox{\linewidth}{!}{
    \begin{tabular}{@{} l cccccccccc @{}}
        \toprule
        \multirow{2.5}{*}{Method}
        & \multicolumn{2}{c}{BUSI}
        & \multicolumn{2}{c}{COVID-QU-Ex}
        & \multicolumn{2}{c}{ISIC-2018}
        & \multicolumn{2}{c}{Kvasir-SEG}
        & \multicolumn{2}{c}{Overall} \\
        \cmidrule(lr){2-3} \cmidrule(lr){4-5} \cmidrule(lr){6-7} \cmidrule(lr){8-9} \cmidrule(lr){10-11}
        & gIoU & cIoU
        & gIoU & cIoU
        & gIoU & cIoU
        & gIoU & cIoU
        & gIoU & cIoU \\
        \midrule
        \rowcolor{gray!15} \multicolumn{11}{l}{\textit{Refering-based Segmentation}} \\
        OVSeg \cite{liang2023open}                  & 4.72 & 4.79 & 17.59 & 16.75 & 26.19 & 24.56 & 13.88 & 13.58 & 13.48 & 18.75 \\
        SAM3 \cite{carion2025sam}   & 17.69 & 0.68 & 0.00 & 0.00 & 1.04 & 1.78 & 0.00 & 0.00 & 5.43 & 1.39 \\
        MedSAM3 \cite{liu2025medsam3}               & \underline{62.93} & \underline{58.97} & \textbf{71.56} & \textbf{75.03} & \textbf{83.03} & \underline{80.43} & \textbf{81.82} & \textbf{82.02} & \textbf{70.99} & \textbf{78.28} \\
        \midrule
        \rowcolor{gray!15} \multicolumn{11}{l}{\textit{Agent-based Segmentation}} \\
        SAM3-Agent \cite{carion2025sam}             & 41.61 & 19.67 & 5.24 & 8.12 & 7.17 & 0.92 & 4.08 & 3.76 & 16.21 & 4.20 \\
        MedSAM3-Agent \cite{liu2025medsam3}         & 58.29 & 55.10 & 46.88 & 46.23 & \underline{82.65} & \textbf{80.88} & 71.89 & 62.84 & 55.71 & \underline{74.33} \\
        \midrule
        \rowcolor{gray!15} \multicolumn{11}{l}{\textit{Reasoning-based Segmentation}} \\
        LISA \cite{lai2024lisa}                     & 41.76 & 48.84 & 62.30 & 63.88 & 71.15 & 56.06 & 57.50 & 56.30 & 55.70 & 56.05 \\
        \textsc{MedREAL} (\textbf{Ours})            & \textbf{69.15} & \textbf{62.52} & \underline{64.51} & \underline{66.92} & 82.21 & 71.74 & \underline{76.35} & \underline{69.73} & \underline{68.49} & 70.47 \\
        \bottomrule
    \end{tabular}}
\end{table*}

\begin{table}[t]
\centering
\caption{Comparison of text generation performance between LISA and our method.}
\renewcommand{\arraystretch}{1.0}
\setlength{\tabcolsep}{2pt}
\resizebox{\linewidth}{!}{
\begin{tabular}{lcccccccc}
\toprule
 & \multicolumn{5}{c}{BLEU} & \multicolumn{3}{c}{ROUGE} \\
\cmidrule(lr){2-6} \cmidrule(lr){7-9}
Method & BLEU & BLEU-1 & BLEU-2 & BLEU-3 & BLEU-4 & ROUGE-1 & ROUGE-2 & ROUGE-L \\
\midrule
LISA\cite{lai2024lisa}  & 24.04 & 56.94 & 29.69 & 18.88 & 13.15 & 59.81 & 33.41 & 48.11 \\
\textsc{MedREAL} (\textbf{Ours}) & \textbf{89.12} & \textbf{98.75} & \textbf{97.74} & \textbf{97.00} & \textbf{96.26} & \textbf{96.69} & \textbf{95.84} & \textbf{96.69} \\
\bottomrule
\end{tabular}}
\label{tab:text_metrics}
\end{table}

\noindent \textbf{Quantitative Segmentation Results.} 
Table \ref{tab:seg_results} presents the quantitative localization performance. When compared to the direct end-to-end reasoning baseline (LISA), \textsc{MedREAL} demonstrates overwhelming superiority, improving the overall gIoU from 55.70\% to 68.49\%. This substantial margin validates that our SARP and R2V modules extract far more precise spatial priors than LISA's holistic \texttt{[SEG]} embedding approach. This advantage is particularly evident in high-noise environments like the BUSI ultrasound dataset, where \textsc{MedREAL} achieves 69.15\% gIoU compared to LISA's 41.76\%, proving our method's robustness against low-contrast boundaries and speckle noise.

It is crucial to correctly contextualize the performance of MedSAM3. While it reaches high metrics (e.g., 70.99\% overall gIoU), this purely structural model benefits heavily from explicit ground-truth category prompts (as defined in Table \ref{tab:referring_phrases}) and extensive pretraining on the original source datasets. However, this label-driven paradigm fails entirely in real-world diagnostic workflows that require multi-step logic rather than explicit naming. This vulnerability is starkly exposed when evaluating MedSAM3-Agent: once forced to autonomously deduce the prompt from complex VQA queries rather than relying on oracle labels, its overall gIoU drastically collapses to 55.71\%. The performance degradation is exceptionally severe on the COVID-QU-Ex dataset (dropping from 71.56\% to 46.88\%), as the decoupled reasoning engine struggles to comprehend complex pathological descriptions, generating inaccurate textual prompts that mislead the segmentation backbone.
In contrast, \textsc{MedREAL} operates entirely on natural language questions without predefined category hints or oracle labels, yet achieves highly competitive performance (68.49\% overall gIoU). By seamlessly bridging the gap between abstract diagnostic reasoning and spatial grounding within a unified latent space, \textsc{MedREAL} establishes a new state-of-the-art for reasoning-driven medical architectures.

\noindent \textbf{Text Generation Quality.} 
Beyond spatial localization, clinically integrated systems must maintain high diagnostic articulation. Table \ref{tab:text_metrics} compares the textual VQA generation quality of \textsc{MedREAL} against LISA. Our framework achieves near-perfect BLEU \cite{papineni2002bleu} and ROUGE \cite{lin2004rouge} scores, outperforming the baseline by massive margins. These results convincingly demonstrate that extracting reasoning semantics for visual guidance does not compromise the MLLM's inherent linguistic capabilities. Instead, our unified architecture successfully harmonizes high-fidelity text generation with precise pixel-level evidence grounding, offering a complete and trustworthy diagnostic response.

\subsection{Ablation Studies}

To rigorously investigate the individual contributions of Box Guidance, the Seg-Anchored Reasoning Pooling (SARP) module, and the Reasoning-to-Visual (R2V) fusion mechanism, we conduct comprehensive ablation experiments. To ensure fair comparison and preserve parameter parity, ablated modules are substituted with equivalent linear projections rather than simply being bypassed. The quantitative impacts across gIoU and cIoU metrics are detailed in Table \ref{tab:ablation}.

\begin{wraptable}{r}{0.61\textwidth}
\vspace{-0.4cm}
\centering
\caption{Ablation study isolating the contributions of bounding-box guidance, SARP, and R2V modules.}
\setlength{\tabcolsep}{4pt}
\renewcommand{\arraystretch}{1.0}
\resizebox{\linewidth}{!}{
\begin{tabular}{cccccc}
\toprule
Box Guide & SARP & R2V & & gIoU (\%) & cIoU (\%) \\
\midrule
\ding{55} & \ding{55} & \ding{55} & & 11.35 & 19.69 \\
$\checkmark$ & \ding{55} & \ding{55} & & 45.02 & 52.84 \\
\midrule
\ding{55} & $\checkmark$ & $\checkmark$ & & 36.24 & 37.74 \\
$\checkmark$ & \ding{55} & $\checkmark$ & & 23.22 & 47.12 \\
$\checkmark$ & $\checkmark$ & \ding{55} & & 45.50 & 46.96 \\
\midrule
\rowcolor{gray!15}
$\checkmark$ & $\checkmark$ & $\checkmark$ & & \textbf{68.49} & \textbf{70.47} \\
\bottomrule
\end{tabular}}
\label{tab:ablation}
\vspace{-0.6cm}
\end{wraptable}

\noindent\textbf{The Necessity of Spatial Priors.}
Removing all three key components degrades performance to a baseline of 11.35\% gIoU, reflecting the inherent difficulty of zero-shot spatial grounding without explicit localization priors. Reintroducing only Box Guidance provides a sharp increase to 45.02\% gIoU. While coarse spatial anchors offer strong regional cues, this configuration still lacks the fine-grained semantic alignment required to delineate precise lesion boundaries.

\noindent\textbf{The Interdependence of SARP and R2V.}
The interplay between reasoning extraction and visual fusion proves critical. When SARP is active but R2V is replaced by a linear mapping, the model achieves 45.50\% gIoU, comparable to the Box Guidance baseline, yet struggles with cIoU. This reveals that while SARP successfully distills concentrated evidence from the reasoning sequence, these semantic features remain bottlenecked without a dedicated fusion mechanism to project them into the visual space. Conversely, enabling R2V without SARP yields a detrimental 23.22\% gIoU; forcefully injecting unrefined, global linguistic representations directly into the visual pipeline introduces severe semantic noise, actively disrupting spatial localization.

\noindent\textbf{Synergistic Integration.}
As shown in Table \ref{tab:ablation}, enabling both SARP and R2V without Box Guidance yields 36.24\% gIoU, demonstrating that reasoning-driven conditioning can partially compensate for missing spatial priors, but remains insufficient on its own. The full \textsc{MedREAL} architecture seamlessly integrates all three components to establish the upper bound (68.49\% gIoU / 70.47\% cIoU). This confirms a highly complementary relationship: Box Guidance anchors the general region, SARP distills the exact diagnostic rationale, and R2V effectively translates this rationale into actionable spatial conditioning.

\begin{wraptable}{r}{0.6\textwidth} 
\centering
\caption{Effect of MLLM scaling on reasoning-driven segmentation performance. Upgrading the backbone directly translates to superior localization.}
\setlength{\tabcolsep}{2pt}
\renewcommand{\arraystretch}{1.4}
\resizebox{\linewidth}{!}{
\begin{tabular}{lcc}
\toprule
Backbone MLLM & gIoU (\%) & cIoU (\%) \\
\midrule
\rowcolor{gray!15}
Qwen3-VL-2B-Instruct & 68.49 & \textbf{70.47} \\
Qwen3-VL-4B-Instruct & \textbf{75.96} & 70.26 \\
\bottomrule
\end{tabular}}
\label{tab:scaling}
\end{wraptable}

\noindent\textbf{Scaling Behavior.}
Finally, we explore the scaling behavior of the underlying multimodal foundation model. As reported in Table \ref{tab:scaling}, upgrading the backbone from Qwen3-VL-2B-Instruct to the 4B-Instruct variant, while freezing the rest of the \textsc{MedREAL} architecture, triggers a substantial gIoU improvement from 68.49\% to 75.96\%. This empirically validates that our framework efficiently leverages the enhanced reasoning capabilities of larger MLLMs, dynamically translating superior linguistic logic into higher-quality evidence masks.

\begin{figure}[t]
    \centering
    \includegraphics[width=\linewidth]{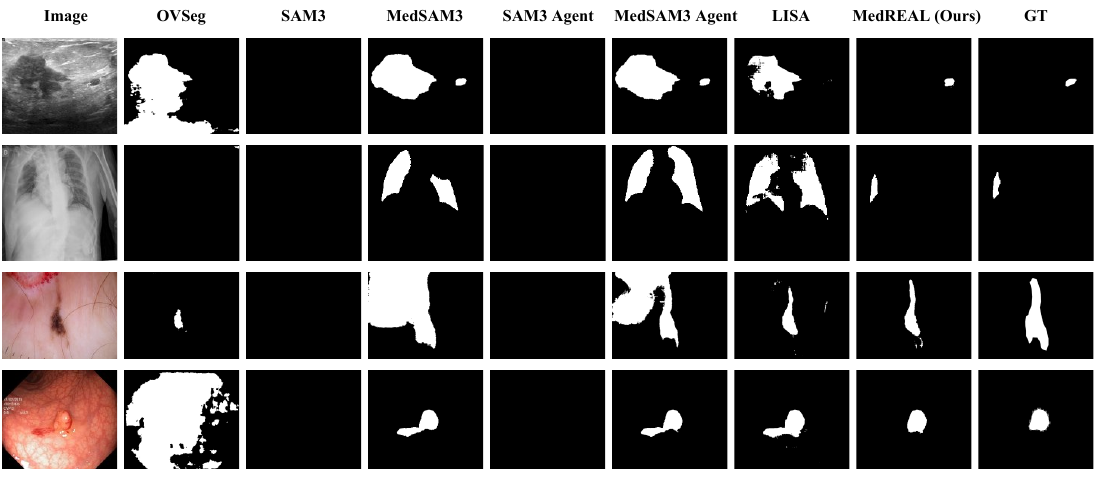}
    \caption{Qualitative comparison with competing methods across four diverse medical imaging modalities (Top to Bottom: BUSI\cite{al2020dataset}, COVID-QU-Ex\cite{tahir2021covid}, ISIC-2018\cite{codella2019skin}, and Kvasir-SEG\cite{jha2019kvasir}). \textsc{MedREAL} consistently produces precise evidence masks that are robust to structural artifacts and highly aligned with the underlying clinical reasoning.}
    \label{fig:qualitative}
    \vspace{0.3cm}
\end{figure}

\begin{figure}[!h]
\centering
\begin{subfigure}[t]{0.49\linewidth}
    \centering
    \includegraphics[width=\linewidth]{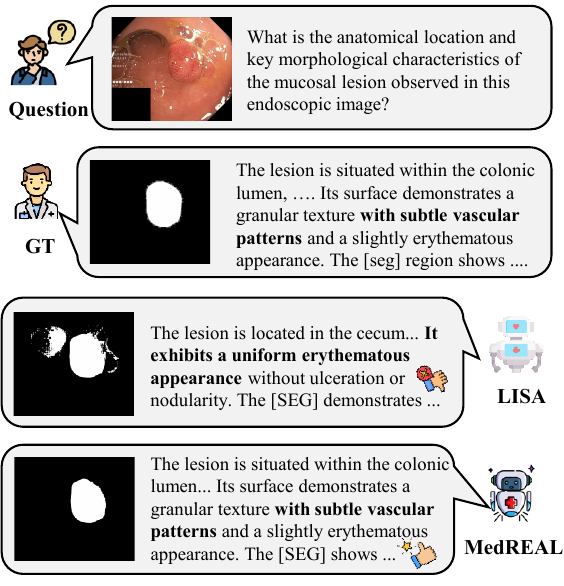}
    \caption{}
    \label{fig:vis2_a}
\end{subfigure}
\hfill
\begin{subfigure}[t]{0.47\linewidth}
    \centering
    \includegraphics[width=\linewidth]{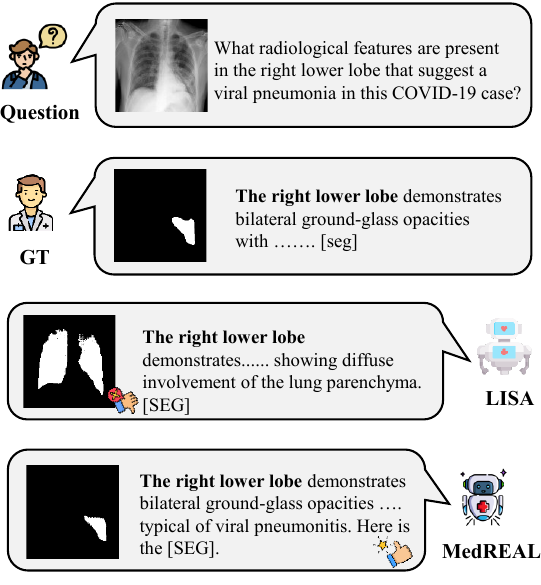}
    \caption{}
    \label{fig:vis2_b}
\end{subfigure}
\caption{Qualitative comparison between \textsc{MedREAL} and LISA. \textsc{MedREAL} demonstrates superior semantic consistency between the generated VQA text and the spatial segmentation mask.}
\label{fig:vis2}
\end{figure}

\subsection{Qualitative Analysis}

\noindent \textbf{Robustness across Modalities.} Figure \ref{fig:qualitative} evaluates segmentation across four challenging clinical scenarios. General-domain models (\eg, OVSeg \cite{liang2023open}, SAM3 \cite{carion2025sam}) fail on low-contrast ultrasound (Row 1) and diffuse X-ray infections (Row 2) due to limited domain knowledge. While MedSAM3 \cite{liu2025medsam3} improves boundary prediction through domain pretraining, its decoupled reasoning makes it sensitive to visual noise, such as endoscopic blood artifacts (Row 4). LISA \cite{lai2024lisa} narrows the semantic gap but relies on a single, global \texttt{[seg]} token. This design limits precise spatial localization, causing LISA to over-segment irregular skin lesion boundaries with small color changes (Row 3). In contrast, \textsc{MedREAL} directly feeds diagnostic reasoning into the visual decoder via SARP and R2V. This rich semantic information ensures robust artifact removal and accurate boundary segmentation across all modalities. Unlike baseline methods that segment the most visually dominant structures, our approach localizes targets guided by underlying clinical logic. Consequently, \textsc{MedREAL} reliably distinguishes true pathological regions from structurally similar healthy tissues.

\noindent \textbf{Reasoning-to-Mask Consistency.} Figure \ref{fig:vis2} analyzes the internal consistency between generated VQA responses and spatial masks. LISA \cite {lai2024lisa} frequently exhibits a serious mismatch: in the endoscopic case (Figure \ref{fig:vis2_a}), its incorrect text prediction of the polyp leads to a false-positive mask, while in the X-ray case (Figure \ref{fig:vis2_b}), it correctly localizes the infection in text but completely fails to locate it spatially. Such unpredictable misalignment greatly reduces user trust in critical medical scenarios, as the visual evidence fails to support the diagnostic text. \textsc{MedREAL} resolves this mismatch. By conditioning the spatial decoding on extracted reasoning features rather than general prompts, our predicted mask serves as an accurate, pixel-level representation of the correct diagnostic text. This reliable alignment turns the MLLM from a black-box predictor into an interpretable and verifiable clinical tool.

\section{Conclusion}

In this paper, we present \textsc{MedREAL}, a unified framework that bridges the semantic gap between high-level diagnostic reasoning and pixel-level spatial grounding in medical multimodal large language models. By introducing the Seg-Anchored Reasoning Pooling (SARP) and Reasoning-to-Visual (R2V) Fusion mechanisms, our approach effectively distills localized semantic evidence from the model's internal reasoning sequence to guide precise mask decoding. Supported by our newly curated MedRAVS-13K benchmark, extensive evaluations across four diverse imaging modalities confirm that \textsc{MedREAL} achieves superior spatial alignment (68.49\% gIoU) compared to existing reasoning-segmentation architectures. Crucially, by strictly coupling linguistic outputs with visual evidence, our method significantly enhances the interpretability and trustworthiness of AI-assisted clinical decision support.

\noindent\textbf{Limitations and Future Work.} 
Currently, \textsc{MedREAL} operates exclusively on 2D imaging modalities and incurs computational overhead due to the autoregressive nature of the MLLM backbone. To address these constraints, our future work focuses on integrating 3D visual encoders into the R2V module to support volumetric data (\eg, CT and MRI) and exploring knowledge distillation to accelerate inference for real-time clinical deployment.

\section{Acknowledgement}
This work was supported by the National Defense Science and Technology Industry Bureau Technology Infrastructure Project (JSZL2024606C001).

%
%
\bibliographystyle{splncs04}
\bibliography{main}
\end{document}